\documentclass{jair}

\usepackage{booktabs}
\usepackage{array}
\usepackage{threeparttable}
\usepackage{mathtools}

\setcopyright{cc}
\acmDOI{10.1613/jair.1.xxxxx}

\JAIRAE{Insert JAIR AE Name}
\JAIRTrack{} 
\acmVolume{4}
\acmArticle{6}
\acmMonth{8}
\acmYear{2026}

\RequirePackage[
  datamodel=acmdatamodel,
  style=acmauthoryear,
  backend=biber,
  giveninits=true,
  uniquename=init
  ]{biblatex}

\begin{document}

\title{Auto-Relational Reasoning}


\author{Ioannis Konstantoulas}
\authornote{Corresponding Author.}
\orcid{0009-0002-8438-6010}
\email{konstantou@ece.upatras.gr}
\affiliation{%
  \institution{Department of Electrical and Computer Engineering, University of Patras}
  \city{Patras}
  \state{Achaia}
  \country{Greece}
}

\author{Dimosthenis Tsimas}
\orcid{0009-0000-7018-7948}
\email{tsimas@ceid.upatras.gr}
\affiliation{%
  \institution{Department of Electrical and Computer Engineering, University of Patras}
  \city{Patras}
  \state{Achaia}
  \country{Greece}
}

\author{Pavlos Peppas}
\authornote{$^{\dagger}$ Deceased. In memoriam.}
\orcid{0000-0002-0008-5505}
\email{pavlos@upatras.gr}
\affiliation{%
  \institution{Department of Electrical and Computer Engineering, University of Patras}
  \city{Patras}
  \state{Achaia}
  \country{Greece}
}

\author{Kyriakos Sgarbas}
\orcid{0000-0002-1797-1343}
\email{sgarbas@upatras.gr}
\affiliation{%
  \institution{Department of Electrical and Computer Engineering, University of Patras}
  \city{Patras}
  \state{Achaia}
  \country{Greece}
}

\renewcommand{\shortauthors}{Konstantoulas, Tsimas, Peppas \& Sgarbas}

\begin{abstract}
{\bf Background \& Objectives:} 
  	In the last decade, Machine learning research has grown rapidly, but large models are reaching their soft limits demonstrating diminishing returns and still lack solid reasoning abilities. These limits could be surpassed through synergistic combination of Machine Learning scalability and rigid reasoning. 
    
    {\bf Methods:}
	In this work, we propose a theoretical framework for reasoning through object-relations in an automated manner integrated with Artificial Neural Networks. We present a formal analysis of the Reasoning, and we show the theory in practice through a paradigm integrating Reasoning and Machine Learning.
    
    {\bf Results:}
	This paradigm is a system that solves Intelligence Quotient problems without any prior knowledge of the problem. Our system achieves $98.03\%$ solving rate corresponding to the top 1\% percentile or 132-144 iq score. This result is only limited by the small size of the model and the processing capabilities of the machine it run on. 
	
    {\bf Conclusions:}
    With the integration of prior knowledge in the system and the expansion of the dataset, the system can be generalized to solve a large category of problems. The functionality of the system inherently favors the solution of such problems in few-shot or zero-shot attempts.
\end{abstract}


\received{25 April 2026}

\maketitle

\section{Introduction}
In recent years, Artificial Intelligence has emerged as one of the most prominent areas of advancement in industry \cite{bharadiya2023rise} and is researched by a plethora of different domains in academia \cite{mariani2023artificial}. This consistent rise and exploitation of the principles of Artificial Intelligence has been mainly fueled by the advancements in Machine Learning seen through the last decade, with the innovation of AlexNet \cite{Krizhevsky2012ImageNetCW}, providing the first glimpse of what Machine Learning was capable of achieving, and less than a decade later, the significant success of Alphafold2 \cite{jumper2021highly} in 2021. A recent prominent result is AlphaGeometry2 \cite{JMLR:v26:25-1654} although using a theorem prover and logical language as additional components. This approach can be classified in a category of Artificial Intelligence called Neuro-Symbolic reasoning, as it utilizes Artificial Neural Networks and discrete reasoning, sometimes called Symbolic Reasoning or Symbolic Artificial Intelligence.

The computational cost of training large models is becoming prohibitive the larger the model becomes, and as early as 2020, there has been literature on the topic \cite{thompson2020computational}. In \cite{thompson2021deep}, the authors claim that the scale of the size that Artificial Neural Network models need to reach grows faster than the increase it provides in capabilities or performance, commonly called diminishing returns. Machine Learning approaches inherently contain limits that should be surpassed through other means, and a proposed solution to this is the Neuro-symbolic approach. Neuro-symbolic is the optimal paradigm because it focuses on integrating the use of both Machine Learning through Artificial Neural Networks and Reasoning through Symbolic Knowledge Representation \cite{garcez2019neural}.

Recent Neuro-symbolic approaches in the literature focus mainly on the learning of constraints or rules through training of models incorporating capabilities of logic learning\cite{skryagin2023scalable}\cite{defresne2026scaling}. In \cite{riegel2020logical} the Logic Neural Networks are proposed where every neuron has a meaning as a component of a formula in a weighted real-valued logic and in \cite{badreddine2022logic} a fully differentiable logical language is introduced. These approaches aim to hybridize discrete reasoning with neural networks, an approach that is very efficient at improving neural network performance. Another way to combine the use of neural networks with the robustness of discrete reasoning is as two separate modules co-operating to solve a problem \cite{trinh2024solving}\cite{JMLR:v26:25-1654}. This co-operative approach leverages the advantages of each system as effectively as possible, but lacks the dynamic learning of rules of recent Neuro-symbolic approaches. This approach can be evolved by integrating the dynamic learning of rules into the discrete reasoning. In this work, we shall introduce a theoretical framework for Auto-Relational Reasoning. We will then use this theory to implement a system as an experimental demonstrator.

Intelligence as a concept and the Reasoning capabilities of humans are measured in literature through metrics such as the Intelligence Quotient (IQ) metric, calculated through many different IQ problems or tests\cite{terman1916measurement}. As Artificial Intelligence popularity grows, the capability of Reasoning of Artificial Intelligence agents has become the subject of research \cite{chollet2019measure}. As such an Artificial Intelligence system that competes in the arena of IQ tests is an interesting research topic.

As a demonstrative experiment to quantify results, we showcase the implementation of an Auto-Relational Reasoning system to solve IQ problems. This system is a Neuro-Symbolic implementation, which realizes the Auto-Relational Reasoning theoretical framework we propose. This approach utilizes a Deep Artificial Neural Network for observations and a symbolic Knowledge Representation in tandem with the Answer Set Programming paradigm to accurately solve problems with significantly higher performance in IQ tests than the median human performance.

\section{Theory}
\label{sec:theory}

The main goal of this approach is to achieve Reasoning with no prior knowledge of the problems, rules, or logic any problem might have. This limitation means that only problems providing enough information to sufficiently constrain the possible solutions can be considered for demonstration. The classic Raven's progressive matrices are chosen for this demonstration, used in IQ test problems. To solve problems generically with logic programming, the problem space must be represented abstractly as a set of objects with traits and relations between those traits.

Let $O = \{o_1, \ldots, o_n\}$ be a set of objects within the problem and let $T = \{t_1, \ldots, t_k\}$ be a set of categorical traits. Let a state $s$ be a tuple $s=\langle t, v\rangle$ where $t\in T$ is a trait and $v$ is a value of a set of values  $V(t) = \{v_1, \ldots, v_m\}$ for each trait $t\in T$. For example an object $o_1$ that is a circle would have $s1=\langle isShape, circle\rangle$ where $isShape \in T$ and $circle \in V(t)$.

Let $C = \{c_1, \ldots, c_n\}$ be a set of categories such that $b$ is a tuple $b=\langle o,c\rangle$ that denotes the relation that object $o$ belongs in $c$. An object can belong to more than one category and each category formally contains a subset of $O$, $\forall c\in C,c\subset O$. A category can be as general as all objects, objects in a repeating position in the grid, objects with a specific shape, or any other abstract category.

We shall denote the argument of an object $o\in O, o \notin C$ and a trait $t\in T$ with value of state $s(t,v)$ as $A(o,s(t,v))$ or more simply $A(o,t,v)$. In the previous example an argument of $o_1$ would be $A(o_1, isShape, circle)$.

An argument can also represent an argument within a category of objects as $A(c,t)$ where $c \in C,c \notin O$. This argument of a category is defined such that $A(c,t)$ categorizes arguments $A(o,t,v)$ for all objects $o \in O$ and $\langle o,c\rangle$.

By $M = \{m_1, \ldots, m_n\}$ we shall denote a set of possible reasoning operators between arguments.
We shall say that $m\in M$ is a reasoning operator between arguments $a_1$ and $a_2$ such that the operation $M(a_1,a_2)$ results in $a_3$.

\[
a_1 = A(o_1, t_1,v_1)
\]
\[
a_2 = A(o_2, t_1,v_2)
\]
\[
m_n(a_1, a_2) =  A(o_3,t_1,v_3) = a_3
\]
\[
for\ o_1\neq o_2 \neq o_3 \in O,\ a_1\neq a_2 \neq a_3,\ t_1 \in T,\ v_1 v_2 v_3 \in V,
\]

Note that in the above operation, only $o_1, o_2, o_3$ and $a_1, a_2, a_3$ are necessarily not equal to other entities of their set. So this operation encompasses most of the possible relations that objects could have in a logic problem setting, even more general than IQ-tests that are just subset of logic problems.

Let each argument-category pair have at least one operator, a relation denoted by the $hasOperator(m,c,a)$ for $m \in M,c\in C$ and $a$ is all $A(o,t,v)$ for $\forall o\in O,\langle o,c\rangle$ or simpler for each $o\in O$ belonging in category $c$. In more common terms, this means each category of objects shall follow at least one operation between them.

The possible operators $m\in M$ are the operators that satisfy the constraints of the relation $hasOperator$. The operators are a function $f:X\xrightarrow{}Y$ such that $\forall x\in X$ there exists exactly one $y\in Y$. A problem is represented using these definitions as predicates and stable model semantics. In this way, the possible solutions of a problem are any answer that satisfies all the constraints set by the relations described above. Any answer that satisfies the constraints is a solution, and an answer that is unsatisfiable under stable model semantics is not a solution. Some problems or some choice of operators for constraining a problem might result in more than 1 answer being a satisfiable stable model through the constraints. If generally more than 1 answer is a solution, then a problem setter has to provide a resolution to this. One resolution is a set of answers where only 1 is satisfiable, and the rest are unsatisfiable. Another resolution is the providence of a method for ranking importance in constraints or, plainly, the flat number of constraints satisfied per answer, rank that answer in importance. This answer set can then be provided with or without importance ranking depending on the resolution chosen.

\section{Methodology}

In this section, we detail the methodology that is employed in the experimental demonstrator used to showcase the utility of Auto-Relational Reasoning in actual systems. The example chosen to test Auto-Relational Reasoning is the solving of IQ problems. The system consists of three main modules, each used in a different phase of the problem-solving process. The first module is the Observation Module that interprets the problem as objects, traits, and categories by using a Convolutional Neural Network (CNN). The second module is the Hierarchical Module that encodes the output of the Observation Module in a knowledge structure to be then translated into knowledge atoms. The final module is the Reasoning Module that leverages Logic Programming rules applied through Answer Set Programming. The problem has thus been transformed into logic atoms, which are then used by a generalizable Answer-Set Programming script that constraints the set of possible answers to only the ones that are solutions.

This methodology is inspired by the work in \cite{kahneman2011thinking} where the Observation Module serves as a System 1 type system for "fast thinking" categorization of object and relations, while the Reasoning Module serves as a System 2 type system for "slow analytical thinking" utilizing the observations of the first system to reduce possible solutions to the problem.

\subsection{Observation module}
The Observation module consists of a Convolutional Neural Network and an encoder that encodes the outputs of the network to a Hierarchical structure of objects and traits. The Observation module can be any computer vision system, or in non-vision problems, any other system that can extract objects, traits, and categories. The CNN architecture described below was chosen since it achieves almost $\sim 0\%$ error. 

\subsubsection{Convolutional Neural Network Architecture}

\begin{figure}[h!]
	\centering
	\includegraphics[width=\textwidth]{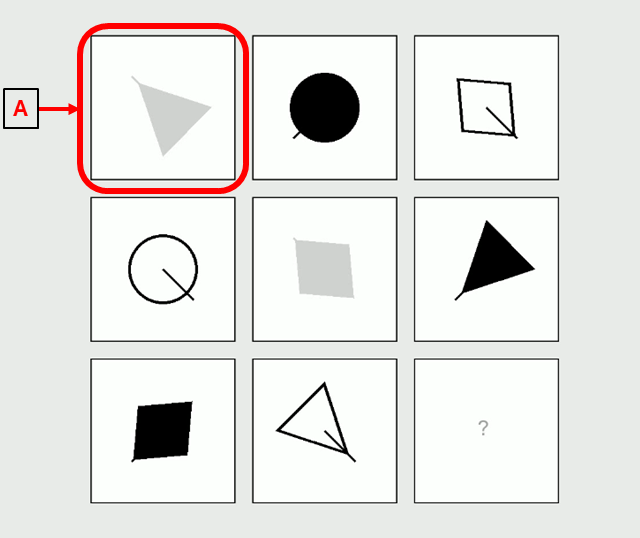}
	\caption{"A" is an example of an input to the CNN.}  
	\label{fig:cnn_input}
\end{figure}

The input of the CNN model is images of shape $250$x$250$ and RGB encoding as $3$ channels. These images are the cells of an IQ problem as shown in Fig. \ref{fig:cnn_input}.

The first layer is a Convolutional layer that contains $32$ filters with a kernel of $3$x$3$ shape, and a ReLU activation function. Its purpose is first-level feature detection, such as edges and textures. This layer is followed by a BatchNormalization and a MaxPooling layer with a $2$x$2$ pool responsible for reducing the size of the feature maps, preventing overfitting. Then, a second Convolutional layer, comprised of $64$ filters with $3$x$3$ kernel size, and a ReLU activation function, is added. The function of this layer is the detection of more complex features. Following that is a Batchnormalization and a MaxPooling layer with the same characteristics as the first ones, again reducing the spatial dimensions while retaining the most prominent features. Subsequently, a Flatten Layer, converting the 2D feature maps into a 1D feature vector, leads to a Fully Connected layer of $1024$ units with a ReLU activation function that is tasked with learning the compounding combinations of input feature patterns in order to make predictions. Lastly, a second Fully Connected layer of $256$ units precedes a set of output heads representing objects. Each output head consists of a presence output with a Sigmoid activation, and trait outputs with Softmax activations, as each trait can have only one value, but many objects can be present at once.

\subsubsection{Training Process}

The model uses Binary Crossentropy loss for the presence heads and Sparse Categorical Crossentropy for each trait head. The Adaptive Moment Estimation (Adam) is utilized as the optimizer. 

The dataset used for training the model includes $400.000$ images of shape $250$x$250$, containing objects that encompass the wide variety of possible object combinations. 

The network was trained using an $80/20$ split for training and validation datasets, respectively. To ensure generalization ability of the model, techniques such as data rotations and value scaling were applied. The learning rate was $0.001$, with the training lasting $20$ epochs and a batch size of $32$. A dynamic learning rate scheduling technique was employed that reduces the learning rate after training loss remains stable for a few epochs. 

The training was conducted on a machine with NVIDIA GeForce RTX $3060$ GPU, $128$ GB RAM, and Ryzen 5600X 6-core $3.7$ GHz CPU.

\subsection{Hierarchical module}
The Hierarchical module consists of an encoder-encoder pair that encodes the outputs of the Convolutional Neural Network firstly into a class hierarchy and then further encodes that hierarchy into logic atoms for the Reasoning module. The first encoder is called the Network encoder, and the second is the Logic Encoder.

The Network Encoder processes observations that the network has output and encodes them into a list of traits for each object. The first task of the encoder is to create a list of objects for each cell of the IQ problem. Traits of an object can be its shape, size, filling, rotation, or anything that could be applied to an object for the purposes of an IQ test. The encoder creates for each object a list of those traits. This process is designed to be abstract and problem-agnostic; as such it can easily be applied to other problems, as long as object-trait relationships suffice for finding a solution. 

The Logic Encoder processes the hierarchical structure created by the Network Encoder and encodes it into logic atoms for the Reasoning module. Each object is encoded as a logic atom, and each object-trait relationship is encoded as an argument. Traits and categories might also be encoded as needed.

\subsection{Reasoning module}

The Reasoning module is an Answer-Set Programming Clingo program. The program has a dynamic and static part. The first part is the dynamic, where atoms and arguments are generated through the Logic Encoder, and the problem is defined in this part. The second part is a generic argument-operator solver, where it is assumed that arguments and objects are tied together through logic or mathematical operations, and the problem is finding which operation is the defining feature of each argument.

The Reasoning Module is operating on a knowledge base of logic atoms constructed by the Hierarchical Module. In this knowledge base, the objects are abstractly represented as a set of traits, states, and relations between those as detailed in Section \ref{sec:theory}.

For this problem, we use a foundation of operators from which all logic operators can be derived \cite{rosen2019discrete}, which make sense for rigid problems such as the Raven's Progressive Matrices. These operators are combined through the use of relations and arguments producing complex logic solving capabilities emergent in the system without any expert intervention. The set of operators used is the Basic Set Operators: Union, Intersection, Symmetric Difference, their negation: Not Union, Not Intersection, Not Symmetric Difference, and a simple first-order series progression $y=a*x$. This reduces the set of acceptable solutions to only answers that satisfy stable model semantics with these operators. The operators allowed in $O$ can be expanded to be functions in such a way that more than one solution sets exist, satisfying the stable model semantics, with some answers satisfying larger constraint sets than others. It can be said that the maximally satisfying answers are the ones commonly accepted as correct, though it is not clear for all problems.

Simple mathematical functions used together with logic can generate more complex functions needed as a operators for more complicated problems. This happens dynamically through the object-trait-argument-value-operator relationship. In \cite{odrzywolek2026all} the authors claim that all mathematical functions can be derived through a single operator. Thus, problems solvable by function application can have a solution derived through the use of a single operator and reasoning through the object relationships, given that the observations are correct.

\section{Results}

\begin{figure}[htp]
	\includegraphics{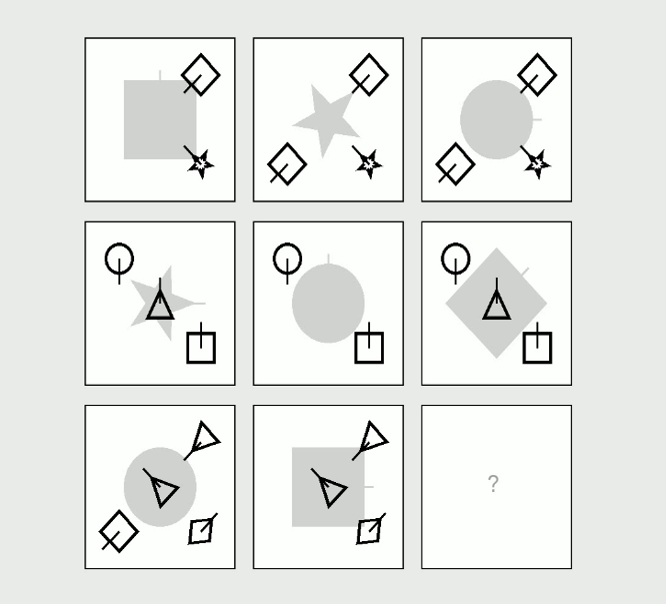}
	\caption{An image of an IQ-test problem given as input to the system.}
	\label{Example1}
\end{figure}

The problems used to measure the performance of the system are IQ-tests with multiple shapes, sizes and different traits, like coloring or rotation. An example of an IQ-test problem is shown in Figure \ref{Example1}.

\subsection{Quantitative Results}

The overall performance of the system is evaluated through the accuracy metric defined as the ratio of accurate solutions produced by the system divided by the total number of IQ questions. An accurate solution is one where the system produces exactly the correct answer. An inaccurate is defined as one where the system either produces two or more answers, one wrong answer, or no correct answer. 

\subsubsection{Observation Module Losses}

In the Observation module there are five core features (i.e. traits), namely the presence of an object, its shape, color, filling, and rotation. For each of these traits we calculate the accuracy and a loss. In the case of presence, we utilize the Binary Crossentropy loss, and for the rest of the traits the Sparse Categorical Crossentropy one. 

In Tables \ref{tab:tab1} and \ref{tab:tab2}, the loss and accuracy of the Observation Neural Network are illustrated.

\begin{table}[ht]
	\setlength\tabcolsep{2pt}
	\centering
	\begin{threeparttable}
		\caption{Object-Trait Loss}
		{\begin{tabular}[t]{l|c|c|c|c|c}
				\toprule
				Object-Traits & \multicolumn{5}{c}{Loss} \\
				\midrule
				&  Presence & Shape & Color & Fill & Rotation \\
				\midrule
				Object 1 & $2.5045\cdot10^{-7}$ & $8.3437\cdot10^{-5}$ & $4.1723\cdot10^{-12}$ & $2.2281\cdot10^{-5}$ & $3.2162\cdot10^{-4}$ \\
				Object 2 & $2.5936\cdot10^{-10}$ & $5.7742\cdot10^{-4}$ & $0.0000$ & $9.6371\cdot10^{-5}$ & $4.3364\cdot10^{-4}$ \\
				Object 3 & $3.8646\cdot10^{-7}$ & $6.0365\cdot10^{-5}$ & $1.0133\cdot10^{-11}$ & $4.4699\cdot10^{-4}$ & $1.5185\cdot10^{-4}$ \\
				Object 4 & $1.4283\cdot10^{-10}$ & $3.0185\cdot10^{-4}$ & $0.0000$ & $4.8211\cdot10^{-5}$ & $0.0012$ \\
				Object 5 & $7.2221\cdot10^{-6}$ & $0.0071$ & $4.3869\cdot10^{-10}$ & $1.9075\cdot10^{-4}$ & $4.9491\cdot10^{-4}$ \\
				Object 6 & $1.3462\cdot10^{-7}$ & $1.2486\cdot10^{-4}$ & $9.8227\cdot10^{-10}$ & $0.0394$ & $0.0035$ \\
				\bottomrule
		\end{tabular}}
		\label{tab:tab1}
	\end{threeparttable}
\end{table}

In Table \ref{tab:tab1}, we observe that the "Presence" trait is the easiest to detect, as it is also the simplest. The most complex trait in terms of detection difficulty is rotation, as it has the highest loss. We can observe that the shape of Object 5 has irregularly high loss compared to the other object shapes. This can be attributed to Object 5 overlapping with Object 6, which occurs when a high number of objects exist in a cell. We can infer similar observations from Table \ref{tab:tab2} as the irregularities in both tables are produced by the same causes.

\begin{table}[ht]
	\setlength\tabcolsep{5pt}
	\centering
	\begin{threeparttable}
		\caption{Object-Trait Accuracy}
		{\begin{tabular}[t]{l|c|c|c|c|c}
				\toprule
				Object-Traits & \multicolumn{5}{c}{Accuracy} \\
				\midrule
				&  Presence & Shape & Color & Fill & Rotation \\
				\midrule
				Object 1 & $1.0000$ & $1.0000$ & $1.0000$ & $1.0000$ & $0.9999$ \\
				Object 2 & $1.0000$ & $0.9998$ & $1.0000$ & $1.0000$ & $0.9998$ \\
				Object 3 & $1.0000$ & $1.0000$ & $1.0000$ & $0.9998$ & $0.9999$ \\
				Object 4 & $1.0000$ & $0.9999$ & $1.0000$ & $1.0000$ & $0.9996$ \\
				Object 5 & $1.0000$ & $0.9978$ & $1.0000$ & $0.9999$ & $0.9998$ \\
				Object 6 & $1.0000$ & $1.0000$ & $1.0000$ & $0.9859$ & $0.9988$ \\
				\bottomrule
		\end{tabular}}
		\label{tab:tab2}
	\end{threeparttable}
\end{table}

\subsubsection{Reasoning Module Accuracy}

To measure the performance of the Reasoning module we need to define what an error would be, and how it can produce wrong results.
The logic problems we deal with in this evaluation are the Raven's progressive matrix problems. These problems are in the context of a $3$x$3$ square of figures given only eight of the figures and being asked to select the ninth from a list of possible answers that only one can satisfy the logic relations implied in the matrix. As such the possible ways the Reasoning module can be mistaken are only two. The first way is a false positive, where the Reasoning module outputs that a false answer satisfies the logic relations implied in the matrix. The second way is a false negative, where the Reasoning module outputs that the correct answer does not satisfy the relations implied in the matrix. Finally the two mistakes can happen at the same time.

For the purposes of measuring the performance of the system, the Reasoning Module is given 8 possible answers where only 1 is correct. Increasing or decreasing the number of answers to choose from changes the performance relatively, as there are more or less opportunities to make false positive mistakes for the Reasoning Module.

The Reasoning module is evaluated in two ways through two datasets. The first dataset is a set of problems, a correct answer, and wrong answers input as atoms in the module. The second dataset is a similarly set of problems as images input through the Observation module, then the Hierarchical module, and finally the Reasoning module. Each dataset contains $40.000$ problems each including $16$ images, $8$ query images, and $8$ answer images.

In Table \ref{tab:tab3}, we calculate the accuracy of Reasoning module using the first dataset. These problems can only be inaccurate due to a mistake made by the problem setter, as the solver always computes correct answers as satisfiable. In this set of problems sometimes two answers might seem mathematically correct to the Reasoning module, though humans might exclude one of the two answers as slightly less correct, through a priori knowledge or common sense.

The accuracy of the overall system is also depicted in Table \ref{tab:tab3}, as Reasoning \& Observation modules using the second dataset. This dataset has the same issue as the first, namely sometimes two answers might seem correct to the Reasoning module, leading to inaccuracies. Another cause of inaccuracies is the Observation module producing incorrect observations, either in the 8 setup query cells or in the 8 answer cells. As expected, this produces worse results than the simpler logic atoms dataset.

\begin{table}[ht]
	\setlength\tabcolsep{2pt}
	\centering
	\begin{threeparttable}
		\caption{Problem Solving Accuracy}
		{\begin{tabular}[t]{l|c|c}
				\toprule
				& Reasoning Only & Reasoning \& Observation \\
				\midrule
				Accuracy & $99.74\%$ & $98.03\%$ \\
				\midrule
				Problems Solved Incorrectly & $103$ & $786$ \\
				\bottomrule
		\end{tabular}}
		\label{tab:tab3}
	\end{threeparttable}
\end{table}

\begin{table}[ht]
	\setlength\tabcolsep{2pt}
	\centering
	\begin{threeparttable}
		\caption{Related Dataset \& Model Accuracy}
		{\begin{tabular}[t]{l|c|c|c}
				\toprule
				Datasets & PGM (WReN) & RAVEN (ResNet+DRT) & ARR (Ours) \\
				\midrule
				Accuracy & $62.6\%$ & $59.56\%$ & $98.03\%$ \\
				\bottomrule
		\end{tabular}}
		\label{tab:tab4}
	\end{threeparttable}
\end{table}

Table \ref{tab:tab4} shows the reported accuracies of related work tested on datasets, namely the PGM dataset \cite{barrett2018measuring} and the RAVEN dataset \cite{zhang2019raven}. This is not a comparison in performance, as all approaches differ in important logical constructions of the problems. Additionally, the related work in the literature is fundamentally different as our method focuses on rigid reasoning using Machine Learning for observations, but the works \cite{barrett2018measuring, zhang2019raven} focus on the possibility of learned reasoning within neural networks. For this reason a true comparison between these works cannot be made, as these works are all unique and only related in the appearance of the experimental demonstration chosen. In \cite{barrett2018measuring, zhang2019raven} the authors' aim is to measure abstract reasoning within neural networks, while our aim is to show the advantages of rigid logic reasoning assisted by neural networks, thus it is unfair to compare results as rigid reasoning is inherently fitted for rigid problems.

The system in total achieves $98.03\%$ solving rate corresponding to the top 1\% percentile or 132-144 iq score depending on the test \cite{bunger2021comparability}. This result far surpasses mean human performance as IQ-tests are standardized to result in about 100 iq, and actual mean results are 102-115 \cite{bunger2021comparability}.

\subsubsection{Ablation and Robustness}

Ablation tests were done to prove the importance of each component. Initially the removal of any atomic knowledge from the Reasoning module produces completely wrong answers at less than 1\% solving rate, with correct solves being completely chance based. The removal or reduction of the layers in the Observation Module result in marginally less accurate results scaling to significantly less accurate the more reductions or removals happen.

As an additional trial, two different cases of reduced dataset quality were attempted. The first trial is the simple reduction on the number of data the network trained on. With reduced data size the performance of the neural networks reduces, with small reductions in size corresponding to small loss in accuracy initial, but the bigger reductions corresponding in significant loss in accuracy. The second trial is the introduction of noise in the dataset, in which the images are induced with variation in color value. These changes affect the early stopping mechanism triggering later making the training of the neural network longer and more demanding while also resulting in a small but not negligent reduction in accuracy.

When a problem includes more possible answers (e.g. 10 instead of 8) the accuracy of the Reasoning module falls by a small margin. The function that can calculate this is the \% chance of error false answer $e$ times the number of answers $n$.

\[
e_n = e*n
\]

\subsection{Qualitative Results}

The main aim of this work is not to focus on the solving of IQ problems, but to automate the reasoning between relations of objects. The example of the IQ problem was utilized, since this problem constrains the possible solutions within the context of the problem, requiring zero a priori knowledge. In this section, we will showcase qualitatively the logical connections that our experimental system determined. The following qualitative results serve as an indication of the capabilities that Auto-Relational Reasoning can provide.  

\subsubsection{Latin Square}

\begin{figure}[htp]
	\includegraphics{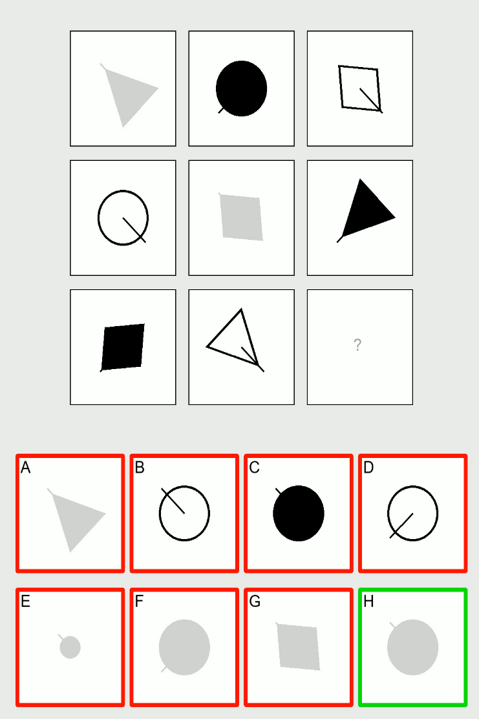}
	\caption{A medium difficulty Latin Square IQ problem.}
	\label{fig:latin_square}
\end{figure}

In Fig. \ref{fig:latin_square}, the Latin Square problem is shown, a classic medium difficulty IQ problem. This is a central problem to solve as it contains multiple rules and connections between rows and columns. The solution to this rule is derived by the system by using the complement of Symmetrical Difference operation. Each shape trait exists only once on each row, each rotation exists only once on each row, and each color/filling combination exists only once. The color/filling and rotation traits are synchronized in this problem, but even if not, the Reasoning module still detects all the rules correctly.    

\subsubsection{The Union Operator}
\label{sec:union}

\begin{figure}[htp]
	\includegraphics{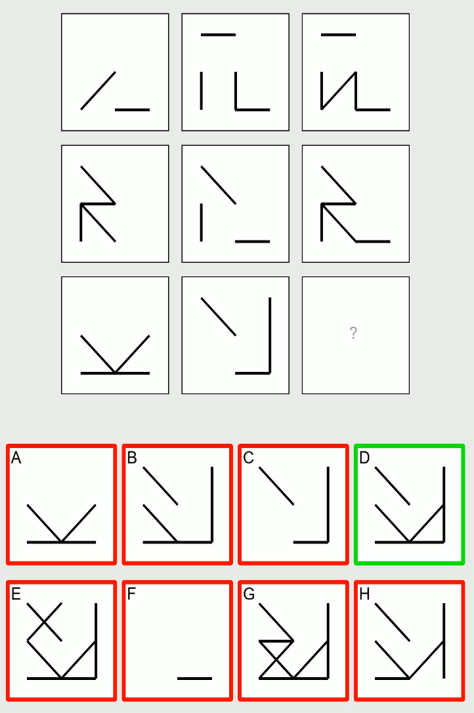}
	\caption{A medium difficulty Union IQ problem.}
	\label{fig:union}
\end{figure}

The Union problem is another classic medium difficulty problem, as seen in Fig. \ref{fig:union}. This is a crucial problem in the progression of an IQ test to teach the solver about pixelwise operations. The system derives the solution to this problem through the elimination of all operators except the valid ones through detecting the rules in the first two rows. One of those is the Union operator, which means only objects that exist in the first or second cell of a row will exist in the third cell. 

\subsubsection{The Symmetrical Difference Operator}

\begin{figure}[htp]
	\includegraphics{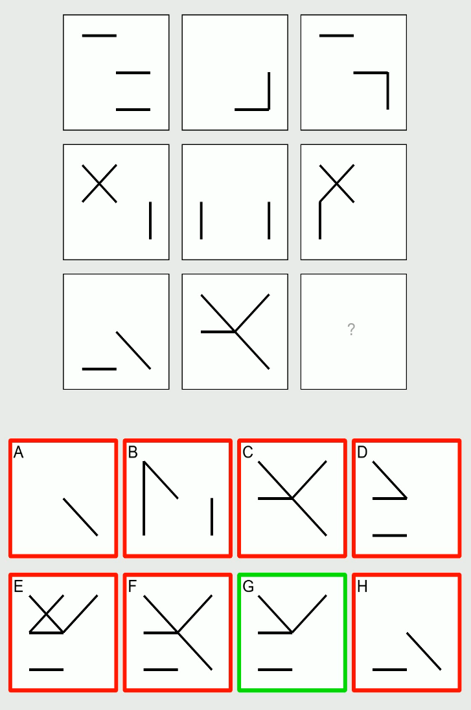}
	\caption{A medium difficulty Symmetrical Difference IQ problem.}
	\label{fig:symmetrical_difference}
\end{figure}

In Fig. \ref{fig:symmetrical_difference}, the Symmetrical Difference problem is depicted, a progression in the difficulty of a problem. The solution to this problem is derived by the Reasoning module in exactly the same manner as the Union problem. All operators except the valid ones are again eliminated by the first two row rules. One of the valid operators is the Symmetrical Difference operator, which dictates if and only if an object exists once between the first and second cells of a row, will it exist in the third cell. Objects that do not exist in any of the first two cells, or exist in both, do not exist in the third cell.    

\subsubsection{Multiple Rule Combinations}

\begin{figure}[htp]
	\includegraphics{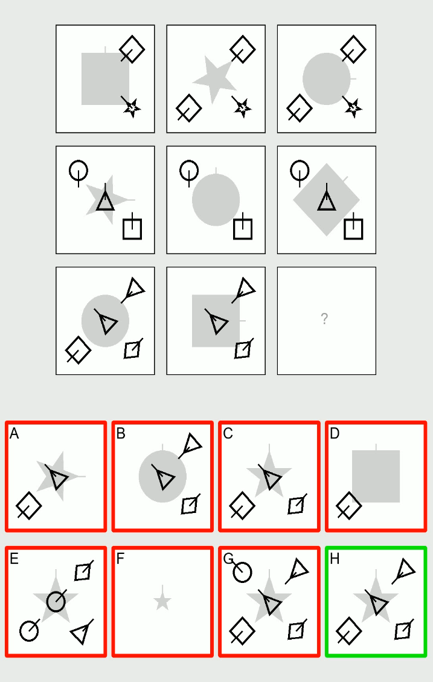}
	\caption{A hard difficulty Multiple Rule IQ problem.}
	\label{fig:multi_rule}
\end{figure}

Hard IQ problems combine multiple rules within a single problem as shown in Fig. \ref{fig:multi_rule}. The problem illustrated in Fig. \ref{fig:multi_rule} is a combination of multiple rules effecting objects at once, based on their categories and traits. In this case, gray objects follow the rule of staying the same for color and filling traits, or it can be said to be the Union or Intersection as no case of gray object changing color or filling is observed. The same objects can be considered a category based on the previous detection, and this category follows the rule of Symmetric Difference Complement for traits shape and rotation. For the foreground objects the rule of Union applies for their presence the same way as detailed in Section \ref{sec:union}.

\section{Discussion}


In this work, we defined a theoretical framework for Auto-Relational Reasoning. This method of reasoning can integrate with machine learning and fully utilize the advancements achieved through years of research in neural networks. Moreover, Auto-Relational Reasoning is based on stable model semantics and grounded in logic, taking advantage of the robust nature of Logic Programming. An experimental implementation was detailed solving Raven's Progressive Matrices problems. This implementation was not tailor-made for the problem, but can be applied to any problem where sufficient information to solve the problem is contained within the problem. 


This experimental demonstrator showcased the capabilities of the framework through an on-the-case implementation of the theory of Auto-Relational Reasoning. The results produced are very promising, with almost perfect accuracy and the main limitation of the implementation being the Artificial Neural Network. The abstraction of the implementation is limited, firstly, by the capabilities of the Observation module. This module can be developed as any computer vision or feature extraction system operating on images or any setup where objects and their traits can be derived from an original source (text, image, sound, video, etc.). Another limitation is the lack of prior knowledge (common sense) in the system and the difficulty of integrating it, as the whole knowledge structure used by the Reasoning module is created dynamically on a per problem basis. This means that we cannot know beforehand what atoms will be created, making some additions to the knowledge base possibly contradictory with runtime dynamically created logic atoms. This limitation is mitigated to an extend through the Observation module, where it is observed that some stochastic systems can approximate properties of prior knowledge (e.g. foreground, background, overlapped, enveloped relations of objects).


For future work, the rational consequent step is the integration of prior knowledge in the system, making it able to solve few-shot or zero-shot problems requiring prior knowledge. An interesting implication of knowledge systems with prior knowledge acquiring new knowledge is the possibility of contradiction and the resolution of it. This resolution is studied in the field of Belief Revision and the integration of Reasoning with Belief Revision is a subject of great academic importance.

\begin{acks}
This work is dedicated to the memory of my late professor, Pavlos Peppas.
\end{acks}

\printbibliography

\appendix

\end{document}